\documentclass[letterpaper, 10 pt, conference]{ieeeconf}  

\IEEEoverridecommandlockouts                              

\usepackage{graphicx}
\usepackage[ruled,vlined]{algorithm2e}
\usepackage{tabularx}
\usepackage{amsmath, amssymb}
\usepackage{subcaption}
\usepackage{cite}

\title{\LARGE \bf
PTDRL: Parameter Tuning using Deep Reinforcement Learning
}

\author{Elias~Goldsztejn$^{1}$, Tal Feiner$^{2}$,
        Ronen Brafman$^{3}$
\thanks{*This work was supported by ISF Grant 1651/19, the Helmsley Charitable Trust through the ABC Robotics Center of Ben-Gurion University, and the Lynn and William Frankel Center for Computer Science}
\thanks{$^{1}$Elias Goldsztejn and $^{3}$Ronen Brafman are with Department of Computer Science at
        Ben Gurion University.
        {\tt\small eliasgol@post.bgu.ac.il, brafman@bgu.ac.il}}%
\thanks{$^{2}$Tal Feiner is with Elbit Systems.
        {\tt\small talfe@post.bgu.ac.il}}%
}

\begin{document}

\maketitle
\thispagestyle{empty}
\pagestyle{empty}

\begin{abstract}

A variety of autonomous navigation algorithms exist that allow robots to move around in a safe and fast manner. However,  many of these algorithms require parameter re-tuning when facing new environments. In this paper, we propose PTDRL, a parameter-tuning strategy that adaptively selects from a fixed set of parameters those that maximize the expected reward for a given navigation system. Our learning strategy can be used for different environments, different platforms, and different user preferences. Specifically, we attend to the problem of social navigation in indoor spaces, using a classical motion planning algorithm as our navigation system and training its parameters to optimize its behavior. Experimental results show that PTDRL can outperform other online parameter-tuning strategies.

\end{abstract}

\section{INTRODUCTION}

Autonomous navigation is a topic of great interest and an active field of research. 
%
Recent  advances in sensing technology and AI made possible major advances in this area. However, mobile autonomous systems still face safety and robustness issues~\cite{Survey_1}, specifically when navigating in crowded indoor spaces.

Classical approaches for collision avoidance 
employ reactive rules for finding safe 
paths~\cite{DWA,ORCA,etb}. While these approaches can be successful in specific domains, they require re-tuning of parameters to accommodate different environments, platforms and situations.

More modern approaches attempt to improve upon these classic algorithms by integrating advances in deep reinforcement
and inverse-reinforcement learning, to capture the complexity of human behaviors, model cooperation between agents and humans, and provide improved environment awareness (e.g.,
\cite{LGM2017,MDD2016,autorl,sacadrl,evita}). 
These approaches 
can lead to better
performance, but because they are often based on end-to-end training, they lack interpretability and transparency, and they can fail badly in unexpected ways on particular inputs 
\cite{intriguing,evaluation}, making it difficult to rely on them. Indeed, autonomous cars have been known to exhibit unexpected catastrophic failures~\cite{NPRCrash,NYTTesla}.

An alternative to using deep RL and deep learning methods for autonomous navigation directly is to try to exploit advances in machine learning to improve classical methods while maintaining their transparency and explainability.
Of particular interest is the approach taken by the APPLD algorithm~\cite{appld} that was used to tune the parameter
of the local planners DWA~\cite{DWA} and ETB ~\cite{etb} in the commonly-used \textit{move\textunderscore base} navigation stack. APPLD was able to considerably improve the algorithm's performance using expert demonstrations and a form of imitation learning. More specifically, first, a human user teleoperates the robot. The trajectories were automatically segmented into $4$ segments types. These contexts correspond to open spaces, spaces with obstacles, narrow corridors, and curves.
Next, a black box optimizer attempts to find parameters for the navigation algorithm, one set of parameters per context, that would yield behavior similar to the human demonstrations. 

\begin{figure}
    \centering
    \includegraphics[width=0.3\textwidth]{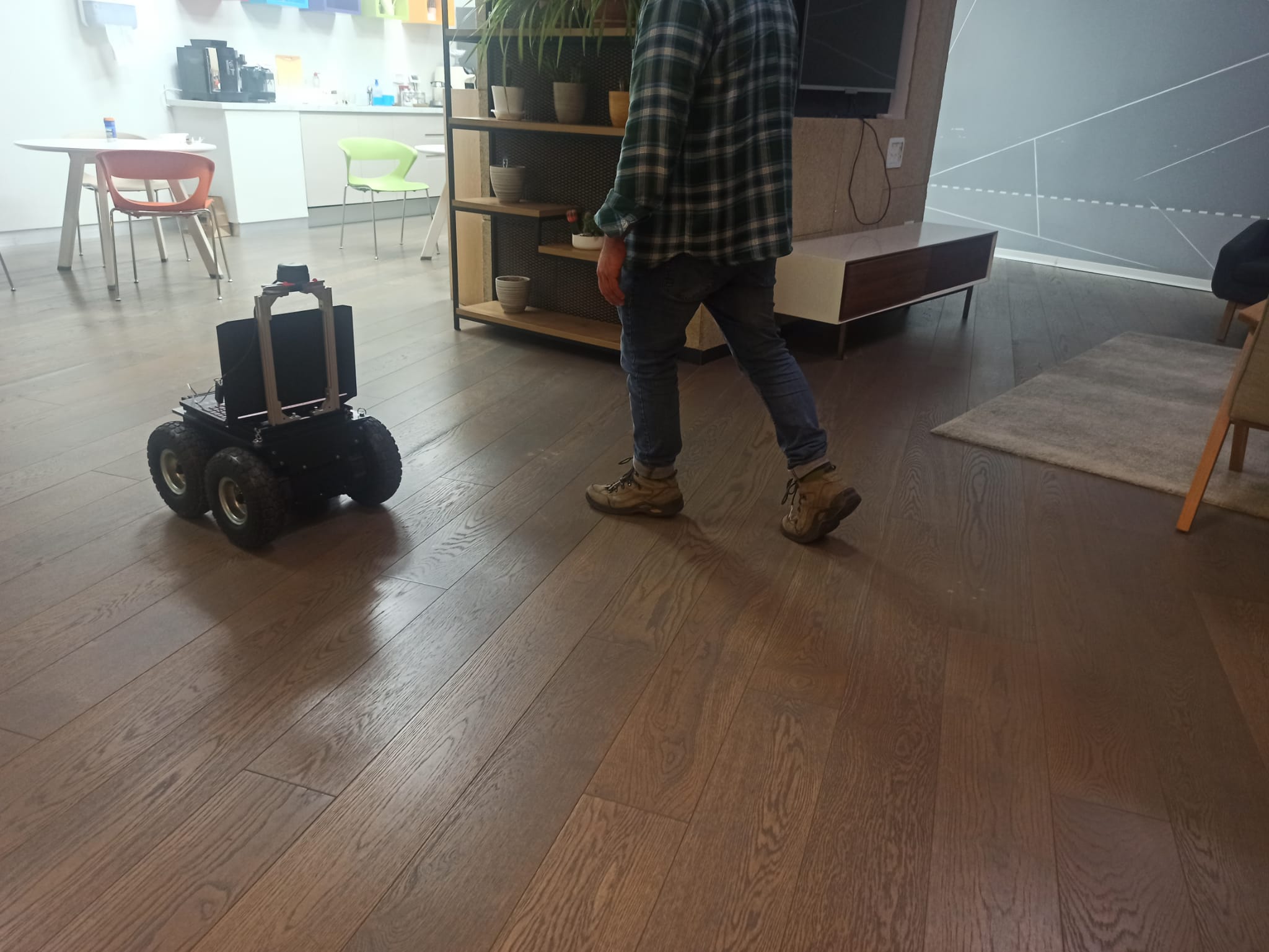}
    \caption{\label{fig:Physical} A robot navigates autonomously among people. According to sensor readings it is able to adapt navigation system parameters to have more desired behaviors.}
\end{figure}

APPLD's ability to considerably improve the algorithm's performance, coupled with the simplicity and transparency of the underlying algorithm, is impressive. Its behavior is easily explainable. However, the method still requires human involvement in the form of new human demonstrations for every new robotic platform, new
environment, and users with different preferences for robot behaviors. Also, the choice of four contexts appears a bit arbitrary.

To address this issue, APPLR~\cite{DBLP:applr}
replaced APPLD's learning by demonstration with reinforcement learning, removing the need for human demonstrations.
By learning policies for selecting parameters
for an existing motion planner instead of directly choosing velocity vectors like previous RL-based methods, APPLR is able to increase exploration safety, improve learning efficiency, 
and allow effective sim-to-real transfer.

However, APPLR has a number of limitations.
Though learning in  parameter space converges faster than in  velocity space, using policy gradient methods to learn in continuous action spaces is very expensive
computationally: APPLR required 6 hours of training on a large cluster of 500CPUs, and it   
mainly optimized speed, placing less attention on  smoothness and safety around obstacles. 
It was also trained on the BARN dataset~\cite{BARN}, which does not include realistic scenarios nor dynamical obstacles.
Finally, it is less simple to explain and understand than APPLD because it continuously changes its parameters in an opaque manner.


The above issues were problematic when we sought to equip a hospital delivery robot with good navigation capabilities. First, we wanted a method with more modest computational needs that could work on a single, decent machine. Second, dynamic obstacles were a major concern in this environment. And while speed was important, so was safety. Finally, we preferred a parameter selection mechanism that is simpler to understand and control. Our main contribution is a novel architecture motivated by APPLR's use of RL for algorithm parameter tuning that requires much more modest computational infrastructure, handles dynamic obstacles well, is fast while maintaining safety, and is easier to understand.

Our solution is conceptually simple: let the user supply a number of parameter sets, and limit the RL algorithm to determine which set to use at each point in time. This is easy to explain and gives the user greater control over the set of allowable behaviors. 
However, to work well, this requires better spatial and temporal context awareness, and hence a more involved learning architecture.

More technically:
$(1)$ We utilize a discrete action reinforcement technique which selects from a set of predefined parameters reducing drastically the search space, thus requiring much less computation. $(2)$ We introduce a simple reward function that captures safer navigation around closer obstacles. $(3)$ Rather than using sensor data to represent the state space used by the RL algorithm, we use a neural architecture that captures historical information and predictions about future obstacle positions, allowing better modeling
of a dynamic environment with moving humans.
$(4)$ By using an encoded latent space, we obtain better generalization, which allow us to use simulation parameters directly on the physical robot 
(Fig.~\ref{fig:Physical}).
$(5)$ We trained our system on a more realistic simulation environment we created that includes humans that move realistically.



Relevant code and videos for this paper are available at {\tt\small https://github.com/eliasgoldsztejn95}

\begin{figure}
    \centering
    \includegraphics[width=0.3\textwidth]{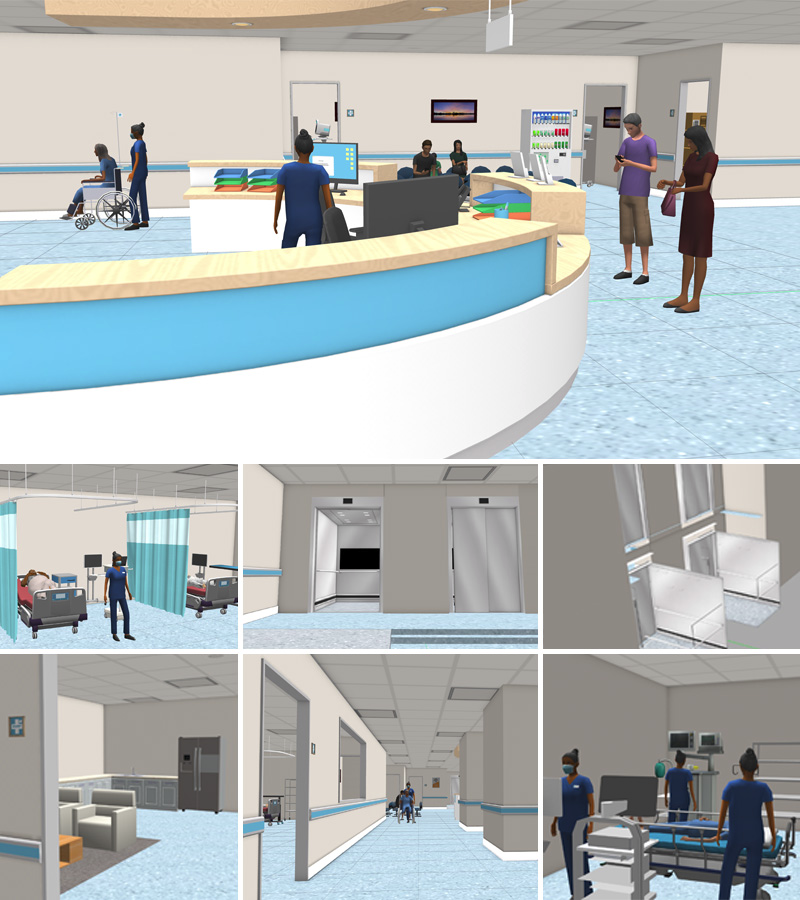}
    \caption{\label{fig:Hospital} A simulated hospital \cite{aws} with humans walking according to social forces \cite{Helbing1995SocialFM}, and a robot travelling through rooms. Collision avoidance, safety and fast movement is needed for the robot to be deemed useful in such an environment.}
\end{figure}

\section{Background and Related Work}

\begin{figure}
    \centering
    \includegraphics[width=0.47\textwidth]{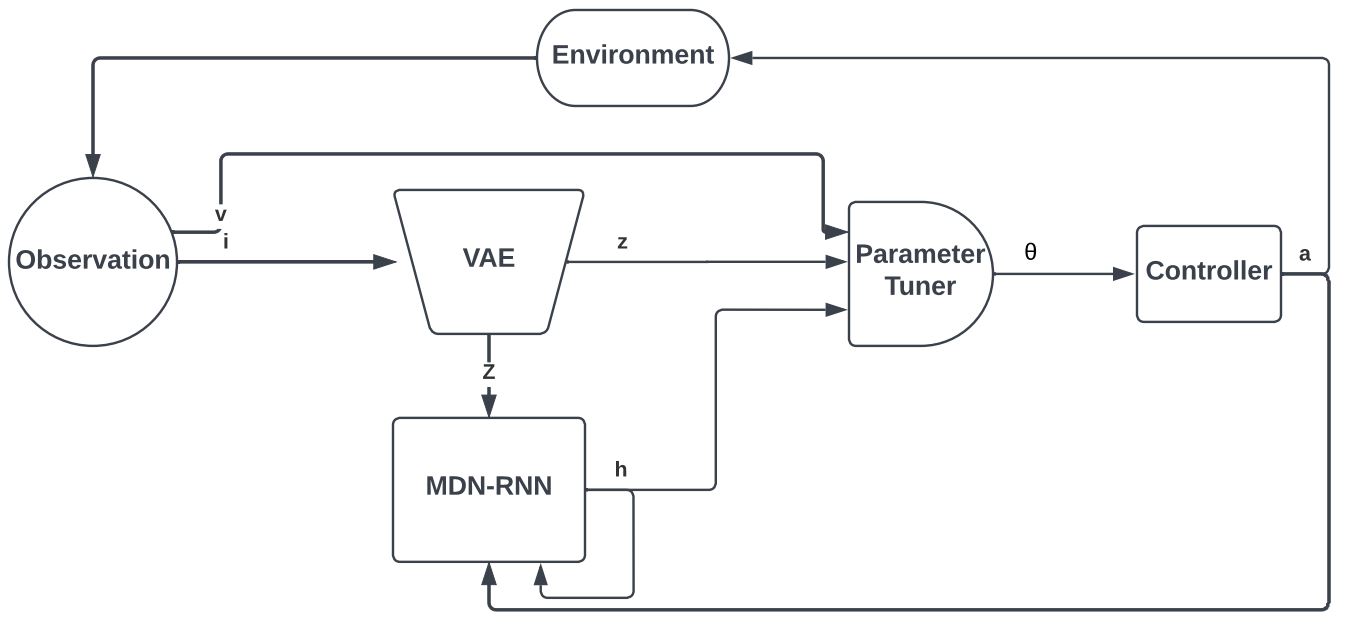}
    \caption{\label{fig:approach2} An observation $[i,v]$ is composed of a birds-eye-view of the robot's cost-map $i$ and velocity $v$. A Variational Auto Encoder encodes a high-dimensional observation into a low-dimensional latent vector $z$. A MDN-RNN (memory module) encodes historical information $z$ to predict future states. A Parameter Tuner outputs $\theta$ which are the parameters for the controller. The controller outputs, $a$, the actions for the robot.}
    \label{approach}
\end{figure}

\subsection{Parameter Tuning}
Most navigation systems have handcrafted parameters that determine the navigation behavior for an entire scenario (a manual for choosing ROS parameters can be found in \cite{manual}).  However an environment may contain different
navigation contexts which require different sets of parameters. For example, a hospital has different contexts like corridors, rooms, waiting areas, etc. and one set of parameters does not perform well in all of them.

An important line of work at U.T.Austin combines parameter tuning with learning approaches, introducing various techniques such as: $(1)$ APPLD which learns from human demonstration, $(2)$ APPLR which learns using reinforcement learning, $(3)$ APPLI~\cite{appli} which learns from human interventions and $(4)$ APPLE~\cite{apple} which learns from evaluative feedback. In this work we focus on APPLD and APPLR.

APPLD approaches parameter tuning in navigation as a contextual problem leveraging machine learning techniques to understand contexts, based on consistent sensory observations and navigation commands. Parameters are optimized to copy an expert by behavioral cloning of human demonstration. A problem with this approach is the dependence on humans, and the problem of replicability among different environments.

APPLR introduces the concept of parameter policy trained
using RL that is capable of choosing sub-optimal actions at one state in order to perform better in the future. It has the advantage of removing the need for human demonstration, while still acting on the parameter space
of a known algorithm, thus retaining advantages like explainability and safety.

APPLR has some practical drawbacks:
it is computationally expensive and its motion is less
smooth than that of APPLD. Its also has some architecture drawbacks: it uses only the present sensor data, which means it both ignores past information and does not attempt to predict the future obstacle states -- this is fine with static obstacles, but problematic with dynamic ones, and it uses raw inputs which have been shown to generalize worse to different situations.

PTDRL views the problem of parameter tuning as a sequential one, where each state is a situational and temporal encoding of cost-maps. This richer state space
provides better contexts for parameter tuning decisions,
and by using an encoder in the process, it has better
potential for generalization. 
PTDRL uses an abstract and dense reward function, that captures both local and global preferences. 

Like
APPLR, it removes the need for human demonstration. Unlike APPLR, it optimizes the cumulative reward choosing from a finite predefined sets of parameters, rather than exploring all the possible combination of continuous parameters. This has the disadvantage of not being able to exploit possible optimal parameters. However, it makes the algorithm much faster to converge,
easier to understand, and, in fact, shows better performance. In the field of parameter tuning -- unlike the field of end-to-end training -- parameters bound the range of possible actions instead of selecting them directly, making a continuous RL potentially redundant. As is shown in this work, a small but distinct spectrum of predefined parameters is easier to learn and performs extremely well. 
 

\begin{figure}
    \centering
    \includegraphics[width=0.47\textwidth]{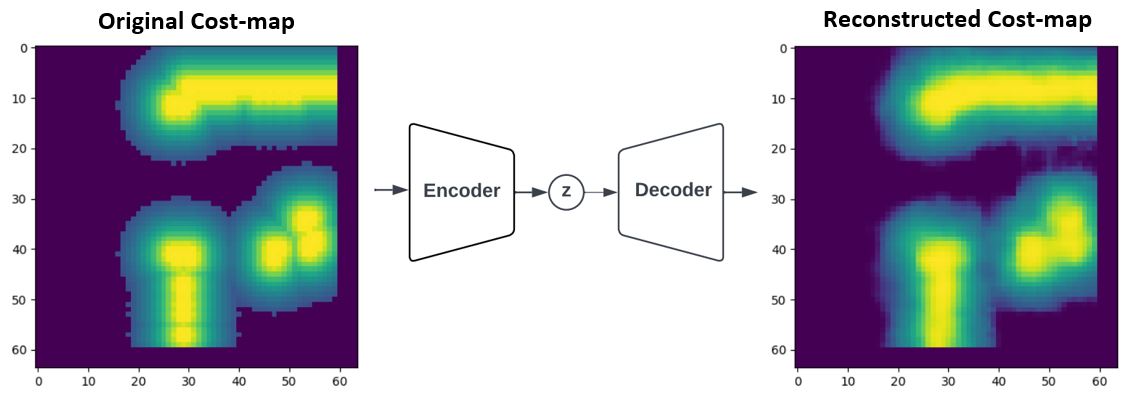}
    \caption{\label{fig:VAE3} Flow diagram of the Variational Autoencoder. The cost-map obtained from a simulation is encoded to $z$. The reconstructed cost-map is shown on the right.}
\end{figure}

\subsection{Spatial and Temporal Context Awareness}
Spatial and temporal awareness are essential for safe and
effective navigation. Spatial awareness enables an agent to form a cognitive model of its physical environment, thereby creating a representation that is useful for navigation. In contrast, temporal awareness is a process that serves as the memory component of navigation, allowing the agent to abstract and retain historical information that can be used to make decisions in the present. 

 APPLD~\cite{appld} demonstrates well the need for context awareness for parameter tuning. In their work, the context is a function of the lidar inputs. They use change-point-detection~\cite{CHAMP} to segment human-guided navigation trajectories into a pre-specified number of contexts. 
 The robot recognizes its current context based on its inputs, and uses parameters suitable for that context.

 APPLR~\cite{DBLP:applr} implicitly uses the raw lidar state (plus a few other parameters) as its context, since this is the input to the policy it generates. While richer than APPLD, this is still, essentially an explicit Markovian state.

The World Models approach (WM)~\cite{worldm} presents a much more elaborate context, or state awareness by combining
a variational auto-encoder (VAE) with an RNN. The encoder learns to generate a compact description of the current input  and  is combined with past inputs using an RNN with a Mixture Density Network to generate predicted states, using which an action choice is made so as to optimize some reward function.

\begin{figure}
    \centering
    \includegraphics[width=0.4\textwidth]{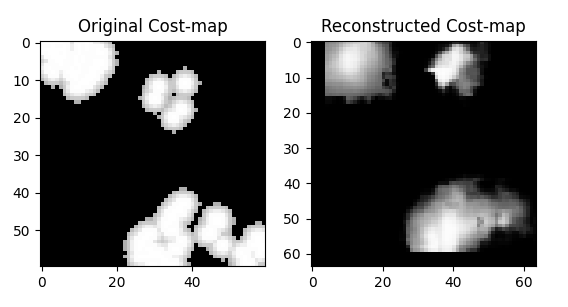}
    \caption{\label{fig:vae_real} Original and reconstructed cost-maps of a physical experiment. The reconstruction captures the main details of the original cost-map, showing that the learnt latent space in the simulation can be used for the real world.}
\end{figure}

We utilize this structure for our navigation algorithm.
Specifically, we use the latent space obtained from a VAE that learns a compressed representation of the observed cost-maps. (Fig.~\ref{fig:VAE3} for simulated environment and Fig.~\ref{fig:vae_real} for the real robot). Then, a predictive model of the future is obtained by combining an RNN with a Mixture Density Network that accounts for the stochastic nature of the environments and the interaction with humans. 
This architecture is shown in Fig.~\ref{fig:approach2}. The latent state is  used as our context.

\subsection{Reinforcement Learning}
Reinforcement Learning deals with the problem of learning to solve sequential decision making problems in a semi-supervised manner. It formulates the decision problem as a Markov decision process, and combines exploration and exploitation to find policies that maximize the cumulative reward over time. It can be helpful in environments where the state-transition dynamics is unknown, and with  advancement in deep learning, it has been able to address policy learning in environments with large or continuous state and action spaces.

We use DDQN \cite{DDQN} which is a model-free, off-policy algorithm that builds on DQN but relies on double Q-learning to avoid the overestimation of action-values. DDQN was shown to perform well on robotic control tasks. It is useful for this setting because it is sample efficient, and is designed to be robust to noisy inputs.
However, our environment is non-Markovian: given dynamic obstacles, information about the past is relevant for predicting the future. Q-learning based methods like DDQN are not suitable for non-Markovian environments. However, by using the richer state generated by WM, we transform the
model into a Markovian one, allowing for the use of DDQN. 


\section{The PTDRL Algorithm}
The following presents PTDRL, a parameter tuning approach that uses deep reinforcement learning to adapt the algorithm's 
hyper-parameters to maximize cumulative rewards.

Abstractly, a navigation system: $C: X \times \Theta \rightarrow A$ maps the state and parameter space to the action space. The state $X$ is represented by the robot sensory inputs and information about the world, such as the cost-map and next way-point. The parameters space $\Theta$ is comprised of optimization parameters of the navigation system, robot constrains, etc. The action space $A$ is a velocity vector (e.g., linear and angular velocity).

Given a specific set of parameters $\theta$ and a state $x$, the navigation system outputs an action $a= C(x,\theta)$. We treat $C$ -- the navigation algorithm -- as a black-box. 
Using deep reinforcement learning techniques we compute
a function: $F: X \rightarrow \Theta$. 
Given $F$, our improved navigation system selects actions in the following way: $C(x;F(x))$. Below we explain  how we
represent and generate the value of the state $x$ and how we learn the function $C$.

\begin{algorithm}
    \SetAlgoLined
    env, rnn, vae   //Environment and network initialization\\
    global variables\\
     \For{episode=1,...,N}{
    obs = env.reset()\\
    h = rnn.initialState()\\
    done = False\\
    cumulativeReward = 0\\
      \While{not done}{
        z = vae.encode(obs)\\
        $\theta$ = parameterController.action([z, h, v])\\
        a = controller($\theta$)\\
        obs, reward, done = env.step(a)\\
        cumulativeReward += reward\\
        h = rnn.forward([a, z, h])\\
      }
     }
    \caption{PTDRL}
\end{algorithm}

\subsection{State Representation}
We assume the robot observes its environment through a Lidar sensor and creates an egocentric 2D image birds-eye-view cost-map.
\footnote{In principle, our method can work with diverse sensor inputs and nothing in it is specific to the use of Lidar. However, all of our experiments are on Lidar-based navigation, and therefore, we make no claim beyond this context.}
Following WM, we use a VAE to learn a compressed abstract representation, $z$, for this input. Thus, $z$ is an embedding of the current cost-map. To incorporate 
historical information we use a residual neural network (RNN) to compress historical information and use the RNN to predict the future. We assume our environment to be stochastic because of the unpredictable trajectories of people. Consequently the RNN is trained to output a probability density function  $p(z)$ as a mixture of Gaussian distributions (MDN). $p(z)$ is the system's prediction of the next state. In our experiments we followed the local planner decision frequency, which was 10Hz, and so the prediction is for 0.1 seconds forward.

The RNN models $P(z_{t+1}|a_t,z_t,h_t)$ where $a_t$ is a concatenation of the action generated by the controller, i.e., the velocity vector, and the inflation radius.
The inflation radius impacts the cost-map, and so knowing the its current value is useful information for predicting the next cost-map. 
$F$ depends on the following components of the state: the current cost-map $i_t$, current velocity of the robot $v_t$ and past action $a_{t-1}$ which form the vector: $[i_t,v_t,a_{t-1}]$. The VAE and MDN-RNN modules use $[i_t,a_{t-1}]$ to generate $z_t$ and $h_t$, and the input into the network is $[z_t,h_t,v_t]$. 
A diagram illustrating the flow process can be seen in Fig. \ref{fig:approach2}.

\begin{table*}[!ht]
    \centering
    \caption{Performance statistics on two sets of 280 simulated test cases. Time and cumulative reward are compared for: Default, APPLD, and PTDRL 4/8. The cumulative reward for each context as well as the total cumulative reward are calculated for each episode. PTDRL 4/8 outperform Default and APPLD in all contexts for both value functions.}
    \label{tab:table_simulation}
    
    \begin{tabularx}{\textwidth}{@{\extracolsep{\fill}} |c||X|X|X|X|}
         \hline
         \multicolumn{5}{|c|}{Reward function 1: $w = 1$} \\
         \hline 
         Context & Default & APPLD & PTDRL 4 & PTDRL 8\\
         \hline
          Time      & $M=81.2s \pm 2.3$       & $M=75.1s \pm 4.9$        &   $\boldsymbol{M=56s \pm 2.1}$     & $M=57s \pm 2.8$      \\
         \hline
         Open       & $M=-103.5 \pm4.0$ & $M=-88.0, \pm3.8$ & $\boldsymbol{M=-84.1 \pm3.3}$& $M=-85.2 \pm3.1$\\
         Corridor   & $M=-30.3 \pm 2.6$ & $M=-40.0 \pm 9.0$ & $M=-30.0 \pm 3.3$& $\boldsymbol{M=-27.9 \pm 2.8}$\\
         Curve      & $M=-25.1 \pm 2.3$ & $M=-19.4 \pm 2.1$ & $M=-19.4 \pm 2.1$& $\boldsymbol{M=-18.8 \pm 2.1}$\\
         Obstacles  & $M=-167.3 \pm 10.3$ & $M=-163.4 \pm 17.3$& $M=-142.2 \pm 9.4$& $\boldsymbol{M=-133.1 \pm 9.1}$ \\
         \hline
         Total  & $M=-326.1 \pm 11.7$ & $M=-310.4 \pm 26.2$ & $M=-274.8 \pm 11.7$& $\boldsymbol{M=-264.8 \pm 10.8}$\\
         \hline
    \end{tabularx}
     
    \vspace{.3cm}
    
    \begin{tabularx}{\textwidth}{@{\extracolsep{\fill}} |c||X|X|X|X|}
         \hline
         \multicolumn{5}{|c|}{Reward function 2: $w = 2$} \\
         \hline
         Context & Default & APPLD & PTDRL 4& PTDRL 8\\
         \hline
          Time     & $M=81.2s \pm 2.3$         & $M=75.1s \pm 4.9$          &  $M=70.1s \pm 1.9$  &   $\boldsymbol{M=69.2s \pm 2.1}$\\
         \hline
         Open      & $M=-118.1 \pm 4.5$ & $M=-99.3 \pm 4.2$ & $\boldsymbol{M=-96.7 \pm 3.5 }$&   $M=-98.4 \pm 4.9 $\\
         Corridor  & $M=-37.1 \pm 3.3$ & $M=-46.3 \pm 8.9$ & $M=-37.6 \pm 3.1 $&   $\boldsymbol{M=-34.2 \pm 3.5 }$\\
         Curve     & $M=-29.9 \pm 2.8$ & $M=-22.7 \pm 2.6$ & $\boldsymbol{M=-22.4 \pm 2.3}$ &   $M=-23.7 \pm 2.6$\\
         Obstacles & $M=-194.2 \pm 11.7$ & $M=-189.2 \pm 19.0$ & $M=-161.5 \pm 10.1 $&   $\boldsymbol{M=-156.6 \pm 10.8 }$\\
         \hline
         Total  & $M=-378.2 \pm 13.4$ & $M=-357.5 \pm 28.3$ & $M=-317.1 \pm 10.1$& $\boldsymbol{M=-311.9 \pm 14.1}$\\
         \hline
    \end{tabularx}

\end{table*}

\begin{figure}
    \centering
    \includegraphics[width=0.4\textwidth]{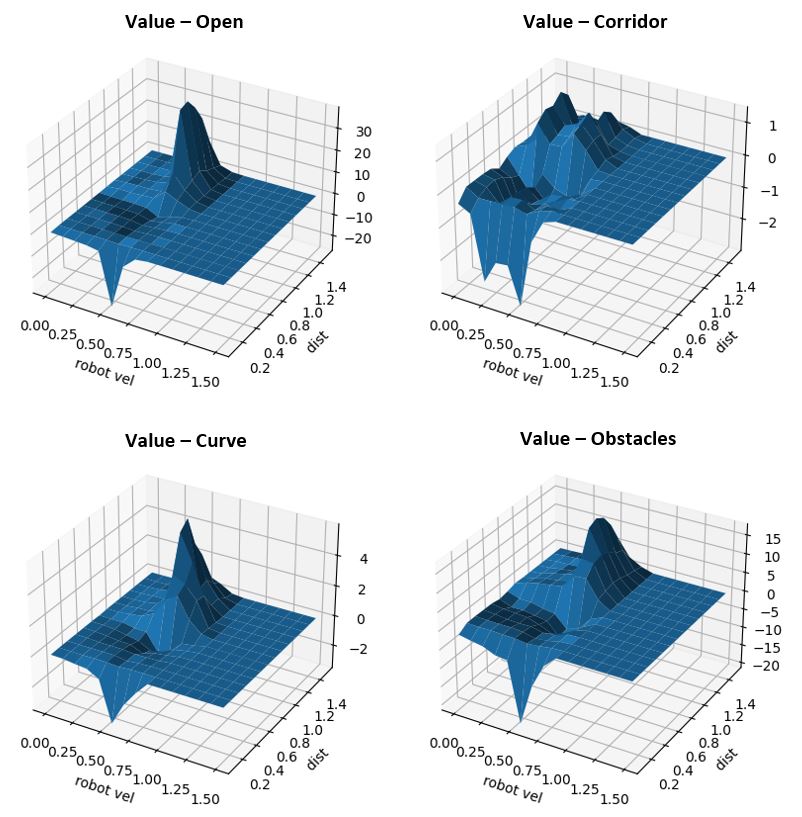}
    \caption{\label{fig:value3} A 3D representation of the value function at different robot velocities and minimal distances. The 3D mesh shows the accumulated values for 70 episodes of a robot running with APPLD parameters.}
\end{figure}

\subsection{Using DRL to Generate $C$}
The RL algorithm attempts to optimize a function
$F: X \rightarrow \Theta$ such that the cumulative discounted reward of the navigation system $C$ is maximized. Intuitively,
in navigation systems our goal is to reach the goal quickly and
safely. The precise definition of the reward function $r$ reflects our preferences regarding these (and possibly other) objectives, and is part of the input to our algorithm. We describe the reward function we used in our experiments in Section~\ref{S:reward}

We chose a DDQN \cite{DDQN} as our DRL algorithm. DDQN uses
a network with a prioritized buffer with three fully connected layers. DDQN is relatively sample efficient and deals well with noise. 
The actions of DDQN are to select at each state one of a discrete set of parameters $\Theta$ that reflect a spectrum of different navigation behaviors. This makes its choices very clear. We expect the user
to provide PTDRL with a set of parameters to choose from. In this work we used the same parameters that APPLD found to be optimal for different contexts in their experiments, and an extra 4 ones for an extended version of PTDRL. 

As we show empirically, an RL algorithm for parameter tuning can work well with a finite set of parameters that induce a expressive enough range of robot behaviors, yet reduces convergence time drastically.


\subsection{Training}
We created a framework in ROS \cite{ros} for reinforcement learning of a robot navigating in an indoor environment (building on the structure of open-AI ROS \cite{openai_ros}), by using a simulation of an hospital \cite{aws} with Gazebo \cite{gazebo}, and modelling humans that move according to social forces \cite{Helbing1995SocialFM} to recreate human behaviour that interacts with the robot. This framework attempts to model
the fact that the movement of humans will take into account the
robot's perceived trajectory. It is a challenging environment
for navigation because of the narrow corridors combined with rapidly moving dynamic obstacles. The DRL algorithm trained on this environment.

For the training of the VAE and MDN-RNN we used synthetic data obtained from the simulated world. We collected about $1$ hour of the robot navigation using different sets of parameters, and collecting cost-maps, local planner velocities and inflation radius. At a testing stage on the physical robot it is observed that the encoding was very accurate on the real world too as can be seen in Fig. \ref{fig:vae_real}

\section{Evaluation}
\label{S:reward}
We compare PTDRL with behavior based on Default parameter values and APPLD's parameters~\cite{appld} for each of the following context defined by APPLD: Open space, Corridor, Curve, and Obstacles. These contexts were classified using a feed-forward neural network classifier, trained on readings of a robot being tele-operated in the simulation environment. We further indirectly compare (due to its high computational requirements) the time metric for the $1^{st}$ reward function with APPLR based on their published results.
Comparisons are conducted in simulation and on a (physical) RoverRobotics Rover Zero 3 robot.

While our method can work with any reward function, we tested it in simulation on two reward functions based on the following parametric reward specification
that incentives high velocities in open environments while penalizing such behavior in more crowded spaces.
$reward = velrob\cdot(-w$ if $mindist < d$ else $1)$ - $maxvelrob$. $velrob$ is the robot's velocity; $maxvelrob$ is its maximal velocity; $mindist$ is its minimal distance from obstacles in meters; $w$ and $d$ are parameters that determine the trade-off between the desire to move quickly and the desire to maintain a safe distance to obstacles. In this work we set $d=0.75$.
Our proposed reward function has a negative value at each time step, thereby driving the algorithm to explore shorter time policies.


\begin{figure}
    \centering
    \includegraphics[width=0.4\textwidth]{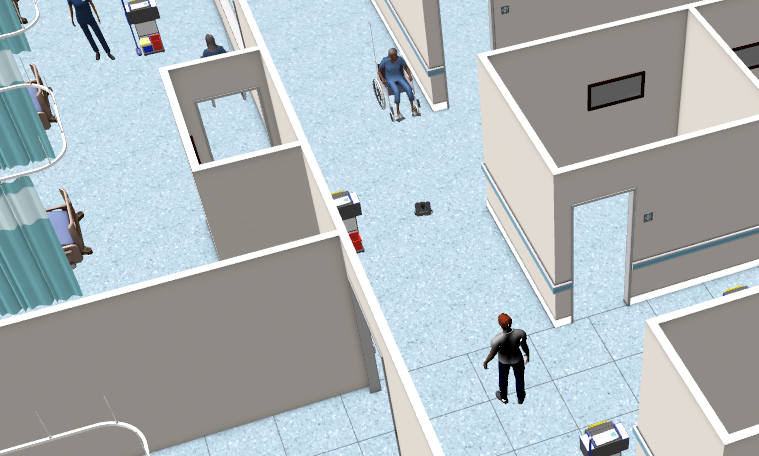}
    \caption{\label{fig:robot_2} A snapshot of the simulation. There is one person moving in a wheelchair in front of the robot and a person behind. PTDRL selects parameters that are aware of the crowded situation.}
\end{figure}

\begin{figure*}[t]
    \centering
    \begin{subfigure}[b]{0.24\textwidth}
        \centering
        \includegraphics[width=\textwidth]{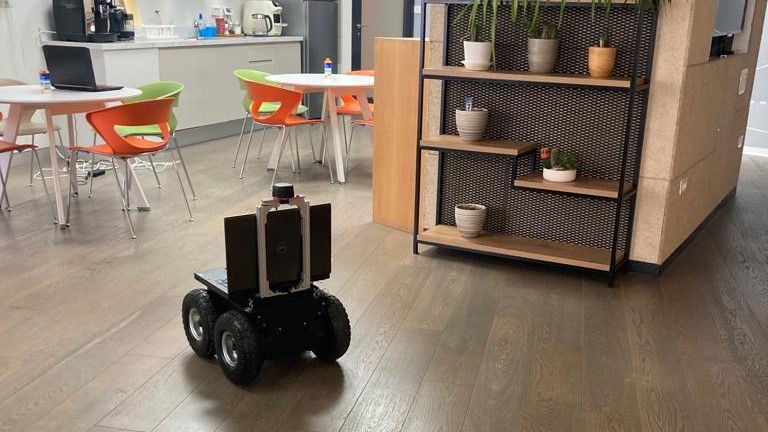}
        \caption{Open}
        \label{fig:fig1}
    \end{subfigure}
    \begin{subfigure}[b]{0.24\textwidth}
        \centering
        \includegraphics[width=\textwidth]{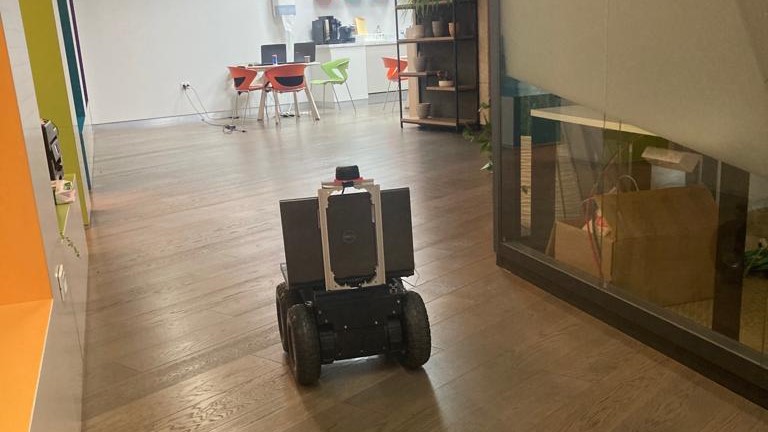}
        \caption{Corridor}
        \label{fig:fig2}
    \end{subfigure}
    \begin{subfigure}[b]{0.24\textwidth}
        \centering
        \includegraphics[width=\textwidth]{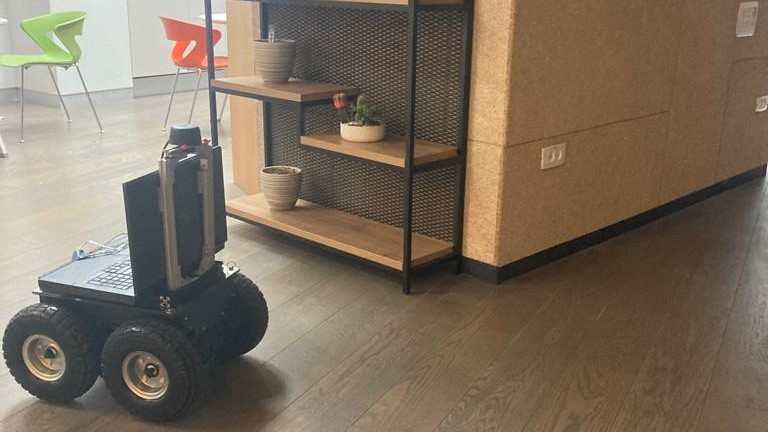}
        \caption{Curve}
        \label{fig:fig3}
    \end{subfigure}
    \begin{subfigure}[b]{0.24\textwidth}
       \centering
        \includegraphics[width=\textwidth]{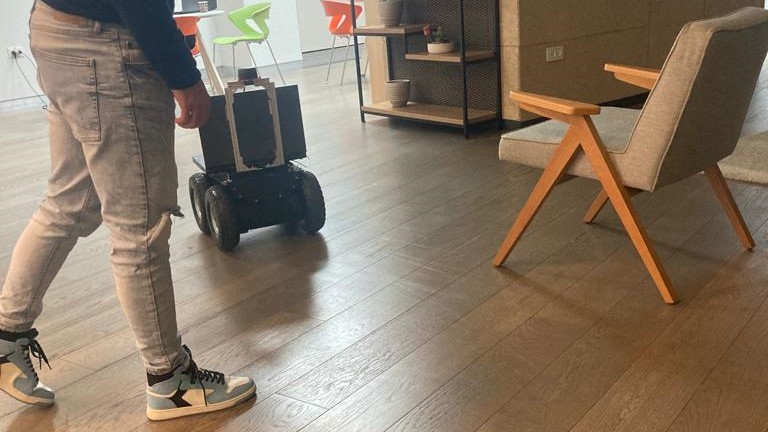}
        \caption{Obstacles}
        \label{fig:fig4}
    \end{subfigure}
    \caption{Rover Zero 3 robot navigating in different contexts.}
    \label{fig:multiple_figures}
\end{figure*}

\begin{table*}[!ht]
    \centering
    \caption{Performance statistics on 40 real test cases. Time and cumulative reward are compared for: APPLD, and PTDRL.}
    \label{tab:table_physical}
    
    \begin{tabularx}{\textwidth}{@{\extracolsep{\fill}} |c||X|X|X|X|}
         \hline
         \multicolumn{5}{|c|}{Reward function 1: $w = 1$} \\
         \hline 
         Context & APPLD time & PTDRL time & APPLD reward & PTDRL reward\\
         \hline
         Open       & $M=11.4s \pm 0.7$ & $\boldsymbol{M=9.4s, \pm0.7}$ & $M=-57.8 \pm5.1$ & $\boldsymbol{M=-49.2, \pm4.8}$\\
         Corridor   &  $M=26 \pm 1.2$ & $\boldsymbol{M=21s, \pm 1.8}$ & $M=-107.4 \pm 7.9$   & $\boldsymbol{M=-72.2, \pm 5.1}$\\
         Curve      &  $M=10.2 \pm 1.0$ & $\boldsymbol{M=8.2s, \pm 0.9}$ & $M=-29.8 \pm 5.8$  & $\boldsymbol{M=-29.2, \pm 5.3}$\\
         Obstacles  &  $\boldsymbol{M=12.6 \pm 0.9}$ & $M=14.2 \pm 0.7$ & $\boldsymbol{M=-57, \pm 8.4}$ & $M=-58.2, \pm 4.8$\\
         \hline
     \end{tabularx}

\end{table*}

\subsection{Computational details}
The networks in this work were implemented in Pytorch \cite{pytorch}. The cost-maps were encoded with a VAE of latent space size of $64$. The hidden layer of the MDNRNN was of size $256$. A DDQN network with prioritized buffer and three fully connected layers of width $290 \times 128 \times 128 \times N_a$, where $N_a$ is the number of parameters in $\Theta$, was trained for about $12$ hours (after $1\cdot 10^6$ steps and roughly $2500$ episodes). The DDQN was tuned to obtain good performance. For both simulation and training, a computer with 
AMD® Ryzen 7 5800x8 CPU, and GeForce. RTX 3060 was used.

\subsection{Simulation environment}
For this work we developed a simulation and training environment using the ROS1 Noetic robot operating system and Gazebo simulation platform. The simulation environment includes a realistic hospital  sourced from \cite{aws}, modeled with humans that move with social forces \cite{Helbing1995SocialFM}. We used two different robots equipped with a Lidar sensor. The navigation system for our robots was \textit{move\textunderscore base}, with the DWA \cite{DWA} classical local planner which plans a robot's trajectory by creating a "dynamic window" of possible velocities and selecting the trajectory that maximizes an objective function, incorporating the robot's dynamic constraints and environment's static obstacles.

The episodes consisted of the robot moving to selected rooms inside the hospital. The cost-maps used for the training input were obtained from the \textit{move\textunderscore base} \textit{localcostmap}, and the velocity of the robot form the \textit{cmd\textunderscore vel} topics. The simulations were sped up about $\times5$ times, taking about $8$ to $14$ seconds for each episode. The training protocol was inspired by \cite{openai_ros} separating the Gazebo, Robot, Task environments, and the training algorithm.

\subsection{Simulation results}
We trained DDQN with $N_a=4$ actions ($4$ sets of parameters). 
These actions controlled the value of the same  parameters which APPLD optimized in their experiments. We used the same sets of parameters for both the DDQN with $N_a=4$ actions and APPLD. We trained another DDQN with $N_a=8$ that included the same APPLD parameters and an extra  set of $4$ parameters. The parameters can be seen in the Appendix.
We compared the Default, APPLD, and PTDRL 4/8 algorithms by running 70 episodes for each algorithm. Each episode consisted of the robot navigating inside the hospital traversing different heterogeneous sectors.
We compare mean values ($M$) with $95\%$ confidence intervals of the following quantities: time of completion of episode, cumulative rewards of contexts, and cumulative reward of episodes. For each episode we saved the immediate rewards of each step, and each step was classified either as: open, corridor, curve, or obstacles - according to a neural network classifier trained previously for segmenting contexts.
We compare both the context specific results as well as the total results over the different trajectories, which involved diverse contexts.
The results are presented in Table~\ref{tab:table_simulation}.
We ran the experiments with two reward functions
setting $\{w=1\}$ and the more
cautious $\{w=2\}$. 

To help understand the impact of different parameter tuning strategies for each context, we generated the graphs in Figure~6. They
show us the robot behavior in different contexts: Given a long simulation run, we record in which context the robot was in at each step, using a segmentating network. For each such state, we
know its velocity and its minimal distance from an obstacle.
The graphs in Figure 6 show us how
often the robot was at each (velocity,distance) point when
in this context, and the correspondent accumulated value of the value function. This enhances user understanding of the impact of different parameter tuning strategies for each context. This helped us create a parameter tuning strategy that is both good locally (for each context) and globally.

PTDRL outperforms Default, and APPLD cumulative rewards in total and for each separate context, showing that PTDRL exhibits both a global and local good performance. It also displays significantly
less variability, as evident from the confidence intervals. PTDRL outperforms APPLR in the time metric too.

For the $1^{st}$ reward function PTDRL outperforms Default and APPLD by a $19 \%$, $15\%$ respectively regarding the total cumulative reward, and $30 \%$, $24 \%$ time-wise, whereas APPLR outperforms APPLD by a $21\%$ in their experiments. There is more incentive for being fast and PTDRL chooses parameters accordingly. However, in contexts like Obstacles and Corridors, PTDRL manages to still choose cautious parameters, thus having good cumulative rewards in those areas.

For the $2^{nd}$ reward function PTDRL outperforms Default and APPLD by a $17 \%$, $13\%$ respectively regarding the total cumulative reward, and $15 \%$, $8 \%$ time-wise. In this reward function there is more incentive for being cautious and
we can see that PTDRL is slower than when using the $1^{st}$ reward function, choosing more cautious parameters in general. Yet, selects less cautious parameters in areas with less obstacles.

\subsection{Hardware experiments}

In this section, PTDRL was implemented on a physical robot (Rover Zero 3) to validate the hypothesis that PTDRL can be successfully applied to a complex real world environment without the need to re-tune any of the neural networks. Furthermore, we compare with APPLD, and show that both algorithms show similar behaviors, and PTDRL is even able to outperform in some contexts.


We performed $10$ experiments for each context,
shown in Fig.~\ref{fig:multiple_figures}, comparing mean time and cumulative reward for each one. The experiments were carried out in an occupied office
building, where we 
simulated contexts similar to APPLD inside the building.

\subsection{Hardware results}
In table \ref{tab:table_physical} the results for the physical experiments can be seen. For the time metric, lower is better. For the reward metric, higher is better. Both APPLD and PTDRL were successful in traversing all contexts in a reasonable amount of time.
While PTDRL outperforms APPLD all contexts except
\textit{Obstacles}, both w.r.t.~time and cumulative reward. In \textit{Obstacles}  APPLD is  slightly better.
Both the APPLD and PTDRL parameters allowed for smooth navigation, and it is important to notice that neither algorithm re-tunes parameters all the time, but rather shows a desired stable behavior.

\subsection{Discussion}
Hardware experiments show that the simulation to reality gap can be bridged well when using a protocol that is able to encode and abstract complex representations of the word into lower dimensional latent spaces, as is the case with WM. As noted earlier, this is seen in Fig.~\ref{fig:vae_real}
that shows successful real world encoding based on 
the network learned in simulation.
The addition of a temporal awareness module, which is the RNN-MDN, further gives navigation systems the ability to reason in time, and react better to dynamic obstacles.

\section{Conclusion and future work}
This work introduces PTDRL, a reinforcement learning strategy for optimizing existing navigation systems via high-level dynamical parameter tuning according to situational and temporal interpretation. Like APPLD and APPLR, PTDRL improves existing navigation systems by learning to adapt parameters dynamically. However, unlike APPLD it learns from simulations instead of needing human demonstration. 
Thanks to its better situational and temporal awareness and its use of encoding, it adapts well to the real world in spite of being trained in simulations. It performs very well, and importantly, it is able to deal well with dynamic obstacles! The method can be easily adapted to different reward functions, environments, robots, and navigation systems; its computational cost are modest; and its behavior is  easy to explain since it essentially
learns to classify each context (state) to one a fixed set of parameters. 

\addtolength{\textheight}{-12cm}   



\section*{APPENDIX}
The table below shows the sets of parameters that we chose from.

\begin{table}[h]
    \caption{PTDRL 4 and APPLD parameters followed by PTDRL 8 extra parameters:
    $max\_vel\_x\,(v),\,max\_vel\_theta\,(w),\,vx\_samples\,(s),$ 
    $ vtheta\_samples\,(t),\,occdist\_scale\,(o),\,path\_distance\_bias\,(p),$
    $ goal\_distance\_bias\,(g),\,inflation\_radius\,(i)$
}
    \label{table_example}
    \begin{tabular}{|c|c|c|c|c|c|c|c|}
        \hline
        v & w & s & t & o & p & g & i\\
        \hline
        1.59 & 0.89 & 12 & 18 & 0.4 & 16 & 7 & 0.42\\
        \hline
        0.8 & 0.73 & 6 & 42 & 0.04 & 32 & 20 & 0.4\\
        \hline
        0.71 & 0.91 & 16 & 53 & 0.55 & 16 & 18 & 0.39\\
        \hline
        0.25 & 1.34 & 8 & 59 & 0.43 & 32 & 20 & 0.40\\
        \hline
        \end{tabular}
        
        \vspace{.3cm}
        
        \begin{tabular}{|c|c|c|c|c|c|c|c|}
        \hline
        0.15 & 1.34 & 8 & 59 & 0.43 & 32 & 20 & 0.15\\
        \hline
        1 & 0.73 & 6 & 42 & 0.04 & 32 & 20 & 0.35\\
        \hline
        0.5 & 0.91 & 16 & 53 & 0.55 & 16 & 18 & 0.3\\
        \hline
        0.2 & 1.2 & 8 & 59 & 0.43 & 32 & 20 & 0.6\\
        \hline
    \end{tabular}
\end{table}






\bibliographystyle{IEEEtran}
\bibliography{IEEEabrv,root}

\end{document}